\documentclass[conference, a4paper]{IEEEtran}
\IEEEoverridecommandlockouts

\ifCLASSINFOpdf
  \usepackage[pdftex]{graphicx}
  \DeclareGraphicsExtensions{.pdf,.jpeg,.png}
\else
\fi

\usepackage[left=1.29cm, right=1.29cm, top=1.9cm, bottom=4.29cm]{geometry}
\usepackage{cite}
\usepackage{comment}
\usepackage{amsmath,amssymb,amsfonts}
\usepackage{algorithmic}
\usepackage{subcaption}
\usepackage{graphicx}
\usepackage{textcomp}
\usepackage{booktabs}
\usepackage{multirow}

\usepackage{optidef}
\usepackage{xcolor}
\usepackage{hyperref}
\usepackage{longtable}
\usepackage{gensymb}
\usepackage[]{algorithm2e}

\usepackage{glossaries}
\usepackage{soul}

 \makeglossaries
\usepackage[font=small]{caption}
\usepackage{enumitem}
\usepackage{anyfontsize}
\usepackage{fancyhdr}
\usepackage{balance}

\makeatletter
\renewcommand{\fnum@figure}{Figure \thefigure}
\makeatother

\title{Aircraft Fuel Flow Modelling with Ageing Effects: From Parametric Corrections to Neural Networks}
\vspace{0.5cm}

\newcommand{\specialcell}[2][c]{%
  \begin{small}%
  \begin{tabular}[#1]{@{}c@{}}%
 #2%
  \end{tabular}%
  \end{small}}
  
\author{%
Gabriel Jarry\IEEEauthorrefmark{1},
Ramon Dalmau\IEEEauthorrefmark{1},
Philippe Very\IEEEauthorrefmark{1},
Junzi Sun\IEEEauthorrefmark{2}
\\[1em]
\begin{tabular}{cc}%
\specialcell{\IEEEauthorrefmark{1}\small EUROCONTROL \\ Aviation Sustainability Unit (ASU) \\
Brétigny-sur-Orge, France} &
  \specialcell{\IEEEauthorrefmark{2}\small Faculty of Aerospace Engineering\\Delft University of Technology\\Delft, the Netherlands} 
\end{tabular}
}

\IEEEaftertitletext{\vspace{-1\baselineskip}}

\begin{document}
\maketitle

\begin{abstract}
Accurate modelling of aircraft fuel-flow is crucial for both operational planning and environmental impact assessment, yet standard parametric models often neglect performance deterioration that occurs as aircraft age. This paper investigates multiple approaches to integrate engine ageing effects into fuel-flow prediction for the Airbus A320-214, using a comprehensive dataset of approximately nineteen thousand Quick Access Recorder flights from nine distinct airframes with varying years in service. We systematically evaluate classical physics-based models, empirical correction coefficients, and data-driven neural network architectures that incorporate age either as an input feature or as an explicit multiplicative bias. Results demonstrate that while baseline models consistently underestimate fuel consumption for older aircraft, the use of age-dependent correction factors and neural models substantially reduces bias and improves prediction accuracy. Nevertheless, limitations arise from the small number of airframes and the lack of detailed maintenance event records, which constrain the representativeness and generalization of age-based corrections. This study emphasizes the importance of accounting for the effects of ageing in parametric and machine learning frameworks to improve the reliability of operational and environmental assessments. The study also highlights the need for more diverse datasets that can capture the complexity of real-world engine deterioration.
\end{abstract}

\section{Introduction}

Aviation is a significant contributor to global greenhouse gas emissions, accounting for roughly 2--3\% of worldwide CO$_2$ output \cite{lee2021contribution}. This impact is expected to grow as air traffic increases, spurring a strong motivation to improve aircraft fuel efficiency and reduce emissions. Accurate fuel flow prediction is essential not only for airline cost and fuel planning but also for evaluating new operational procedures and technologies aimed at reducing fuel burn. However, achieving high accuracy in fuel flow estimation is challenging due to the complex interplay of aircraft performance, operations, and environmental factors.

The aviation community has developed standardized performance models to estimate fuel consumption. A prominent example is EUROCONTROL’s Base of Aircraft Data (BADA) model \cite{nuic2010bada}, which provides generic fuel flow and performance profiles for various aircraft types. Such classical models are parametric, built on physics-based principles (e.g., energy balance and specific fuel consumption curves) and calibrated to manufacturer data. They enable quick fuel burn estimates for trajectory simulations and fleet-level analyses. Nonetheless, these models have known limitations: they assume “nominal” aircraft performance and do not explicitly account for aircraft-to-aircraft variability or changes over an aircraft’s service life. In practice, factors like engine degradation or maintenance actions can cause an aircraft’s fuel efficiency to deteriorate as it ages. For instance, studies have indicated that an aged engine can require several percent more fuel than a fresh engine for the same flight mission \cite{arrieta2024engine}. Standard models like BADA, which lack a representation of ageing effects, may therefore systematically underestimate fuel burn for older airframes. This gap is increasingly important as operators consider longevity and retrofit programs: without incorporating ageing, long-term emission forecasts and fuel planning for older fleets may be optimistic.

Integrating aircraft ageing or deterioration effects into fuel flow models is crucial for improving prediction fidelity. If we can quantify how fuel flow trends change over an aircraft’s lifecycle, we can correct baseline models to better reflect real-world performance. A more accurate fuel burn model would enhance environmental impact assessments by accounting for fleet age mix. Furthermore, the industry push for greener aviation (through measures like optimized flight trajectories, new engine technologies, and alternative fuels) requires robust benchmarks; these benchmarks should consider that a 20-year-old aircraft will not perform like it did when new. In summary, there is a pressing motivation to improve fuel flow prediction by bridging the gap between idealized models and the reality of ageing aircraft.

In this paper, we present a comprehensive investigation into methods for integrating engine deterioration and aircraft ageing effects into fuel flow modeling. Using a rich dataset of nearly nineteen thousand Quick Access Recorder (QAR) flights from nine Airbus A320-214 airframes spanning a wide range of service ages, we systematically evaluate classical parametric models, empirical age-dependent corrections, and modern machine learning architectures. Our approach includes augmenting neural networks with explicit age information either as an input feature or through structured inductive biases that encode expected ageing behaviour.

We demonstrate that standard baseline models, such as EUROCONTROL’s BADA, tend to underestimate fuel consumption as aircraft age, reflecting their assumption of nominal, time-invariant performance. Incorporating ageing effects significantly reduces bias and improves predictive accuracy, although challenges remain due to limited maintenance data and the relatively small number of individual aircraft represented. The contributions of this work are twofold: (i) quantifying the impact of ageing on fuel flow prediction accuracy across diverse modelling paradigms, and (ii) proposing novel hybrid approaches that embed ageing effects within data-driven frameworks to enhance robustness and interpretability.

The remainder of the paper is structured as follows: Section \ref{sec:literrature} reviews relevant literature on fuel flow modelling and ageing effects; Section \ref{sec:dataset} describes the dataset and its characteristics; Section  \ref{sec:method} details the methodological approaches; Section  \ref{sec:result} presents results and comparative analysis; Section  \ref{sec:discussion} discusses limitations and implications; and Section  \ref{sec:conclusions} concludes with perspectives for future work.

\section{State of the Art}
\label{sec:literrature}

Before delving into our contribution, it is essential to situate our work within the broader landscape of fuel burn modelling. The estimation of aircraft fuel consumption has been the subject of extensive research over several decades, evolving from purely physics-based formulations to sophisticated data-driven and hybrid approaches. In this section, we review the main strands of the literature, highlighting how classical parametric models, machine-learning methods, and physics-informed approaches have each contributed to the state of the art. We also examine studies on engine deterioration and ageing effects, which provide crucial context for understanding long-term performance trends. This overview lays the foundation for identifying the remaining research gaps that our study aims to address.

\subsection{Classical Parametric Models}
Classical fuel burn estimation relies on simplified physics-based formulations that capture the primary energy balance and specific fuel consumption (SFC) characteristics of turbofan engines. The EUROCONTROL Base of Aircraft Data (BADA) model exemplifies this category, offering phase-specific fuel flow equations parametrized by manufacturer-provided coefficients \cite{nuic2010bada, nuic2010user}. While BADA underpins many trajectory simulators and global emissions inventories due to its computational efficiency and transparency, it assumes “nominal” aircraft performance, neglecting within-type variations and operational factors beyond flight phase and mass. Extensions such as the FEAT reduced-order model illustrate how BADA outputs can be combined with regression techniques to estimate fleet-level fuel use within 5\% of reported values \cite{seymour2020fuel}. It provides an empirical formula that can be applied to adjust BADA fuel flow for older aircraft:
\[
\text{Seymour\_coeff}(\text{age}) = \frac{100}{100 - 1.28 \times \log(\text{age} + 1)}.
\]

OpenAP further democratizes parametric modelling by providing an open-source, research-focused implementation of energy-balance equations \cite{sun2020openap}, and recent aerodynamic-theory methods have proposed an alternate aircraft open-source modelling for cruise and other phases  \cite{poll2021estimation,poll2021estimation2}. Despite these advances, classical models typically treat aircraft performance as static over time, leading to systematic biases when additional factors like ageing and deterioration effects alter real-world fuel flow.

\subsection{Pure Data-Driven Models}
Driven by the proliferation of flight-recorded and simulated data, purely statistical methods have been developed to learn fuel burn relationships directly from observations. Gaussian process regression has been applied to operational data to refine emissions inventories, capturing nonlinear dependencies with quantified uncertainty \cite{chati2017gaussian}. Tree-based ensemble methods, such as gradient boosting and random forests, further improved per-flight fuel estimates by integrating sensor-derived features \cite{chati2016statistical}. Support vector machines and component-level simulations have been explored to diagnose engine deterioration through fuel flow anomalies \cite{weili2014performance,zhao2014diagnosis}. Neural network architectures, including radial basis function networks and feed-forward multilayer perceptrons, demonstrated strong performance across flight phases, outperforming parametric baselines on full-flight sensor datasets \cite{baklacioglu2021predicting,jarry2020approach, jarry2021toward, jarrytowards}. Time-series models such as long short-term memory (LSTM) networks have been tailored to predict contingency fuel requirements and emissions profiles, showing promise for real-time applications \cite{li2021study,metlek2023new}. However, data-driven methods often require extensive, high-quality training data, and their black-box nature can obscure failure modes when encountering out-of-distribution operating regimes or ageing-related drift.

Scaling neural networks to millions of flight-record samples has enabled the capture of complex, non-linear trends in fuel flow predictions. Jarry et al.\ trained deep architectures on thousands of Automatic Dependent Surveillance-Broadcast (ADS-B) derived flights spanning over 100 aircraft types, incorporating domain generalization techniques, such as stochastic perturbations of performance parameters to reduce error on unseen types \cite{jarry2024generalization}. Huang et al.\ demonstrated that publicly available ADS-B data can act as a surrogate for proprietary QAR data when estimating fuel consumption, by combining classification trees with neural regressors \cite{qu2019modelling}. Despite robust accuracy, deep models typically define each aircraft by static type identifiers, and variability in service life or individual maintenance histories is not accounted for in these studies.

\subsection{Hybrid and Physics-Informed Methods}
Hybrid approaches seek to synergize the interpretability of physics-based models with the flexibility of machine learning. One strategy involves data-driven calibration of BADA coefficients against QAR measurements: Uzun et al. showed that per-aircraft thrust and SFC coefficient tuning reduces fuel prediction errors from around 3\% down to under 1\% for individual flights \cite{uzun2021physics}. Neural networks augmented with physics-based loss functions during training have further improved extrapolation robustness to parameter variations \cite{baklacioglu2016modeling,uzun2021physics}. These hybrid frameworks enhance accuracy and interpretability but still generally assume time-invariant performance parameters.

\subsection{Engine Deterioration and Ageing Effects}
Empirical analyses quantify the impact of component wear on fuel efficiency. In 1980, Mehalic and Ziemianski first documented increases in SFC due to compressor fouling and turbine erosion \cite{mehalic1980performance}. Decades later, Arrieta et al. projected a 4.5\% fuel penalty over CF34-series service life via detailed cycle modelling \cite{arrieta2024engine}. Naeem et al. developed deterioration cycle models showing that in-service degradation shifts engine operating points, requiring higher fuel flow to sustain performance \cite{naeem1998implications,naeem2008impacts,naeem2012implications}. Advanced anomaly detection using neural networks and fuzzy logic achieves high accuracy in identifying early-stage degradation \cite{kurt2023prediction,amrutha2019aircraft}. Navaratne et al. incorporated degraded-engine performance into trajectory optimization, revealing that using clean-engine optimized paths on aged engines incurs significant fuel and emissions penalties \cite{navaratne2019impact}. Another research proposed a logarithmic formula based on aircraft age that computes a deterioration coefficient to multiply the baseline fuel consumption, thereby accounting for engine efficiency loss over time \cite{seymour2020fuel}. Environmental factors, such as reduced air density at cruise, exacerbate degradation impacts, though alternative fuels may partially offset these losses due to their higher heating values \cite{koh2018performance,koh2018computational}.

\subsection{Research Gap}
While classical, statistical, deep learning, and hybrid methods provide a comprehensive toolkit for fuel flow estimation, the explicit incorporation of aircraft ageing remains limited. Most models presume static performance parameters, neglecting within-type and within-aircraft degradation trends that evolve over service life. This omission can introduce systematic over or underestimation of fuel burn, compromising cost planning, emissions forecasting, and operational decision support. Our work addresses this gap by systematically analysing QAR data across aircraft lifecycles and developing integrated modelling strategies that explicitly account for age-related performance drift.

\section{Dataset}
\label{sec:dataset}

\begin{figure*}[ht]
    \centering
    \includegraphics[width=0.7\textwidth]{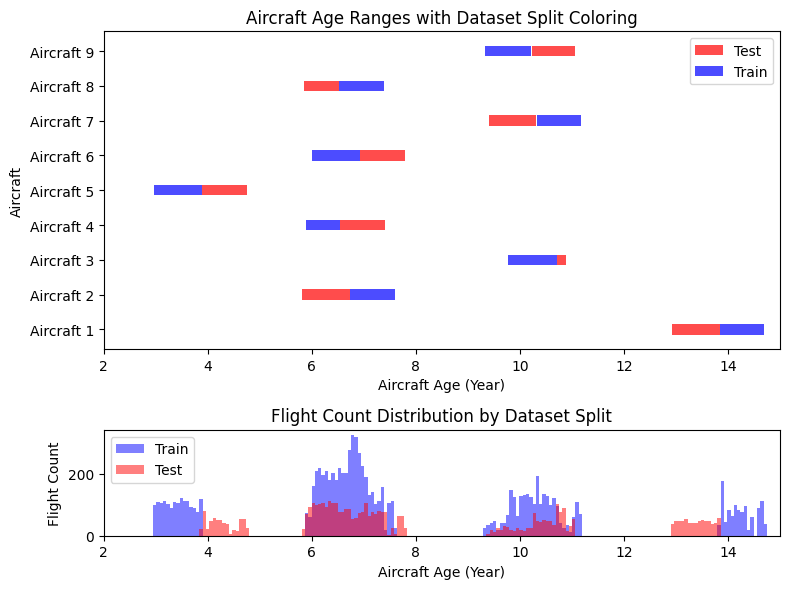}
    \caption{Distribution of aircraft ages in the dataset, illustrating the range of service years covered by the nine A320-214 airframes.}
    \label{fig:aircraft_distrib}
\end{figure*}

The dataset comprises approximately 19,000 Quick Access Recorder (QAR) flight records from a single aircraft type, the Airbus A320-214, covering nine distinct airframes over a period of around two years. Although detailed information on engine cleaning cycles is unavailable, the dataset includes the age of each aircraft measured from its first commercial flight. Importantly, the dataset does not span the full service life of each airframe: it contains about two years of flights, during which the aircraft ages at the time of recording range from 3 to 14 years. This age distribution across the fleet is visualized in Figure~\ref{fig:aircraft_distrib}, highlighting the representation of both relatively new and mature aircraft within the dataset. The dataset is split for each aircraft based on the date (half of the overall calendar period) and represents a split of around 2/3 in train and 1/3 in test. This design choice and associated limitations are discussed in the Section \ref{sec:discussion}.

Each flight record contains 4-second-rate measurements of various operational and environmental parameters, enabling a granular analysis of fuel flow behaviour over time. The dataset’s diversity in aircraft age and operational conditions offers a unique opportunity to explore the impact of ageing on fuel consumption and to calibrate both parametric and data-driven models accordingly. Table \ref{tb:aircraft_characteristics} summarizes the set of input features extracted from QAR data and metadata, along with the target variable used for model training.

\begin{table}[h!]
\centering
\caption{Main model variables: QAR sensors, metadata, and target.}
\label{tb:aircraft_characteristics}
\begin{tabular}{ll|ll}
\toprule
\textbf{Variable} & \textbf{Unit} & \textbf{Variable} & \textbf{Unit} \\
\midrule
Pressure altitude     & ft     & True airspeed (TAS)   & kt \\
$\Delta$TAS / $\Delta t$ & kt/s & Vertical speed         & ft/min \\
Ground speed          & kt     & Mass                   & kg \\
Aircraft age          & yr     & Static air temperature & K \\
Fuel flow (target)    & kg/h   &                        &    \\
\bottomrule
\end{tabular}
\end{table}

\vspace{-1mm} 

\section{Methodology}
\label{sec:method}

This study investigates multiple approaches to incorporate aircraft ageing effects into fuel flow modelling, benchmarking them against two established baseline models that do not explicitly account for engine deterioration or age-related performance changes. The models are trained jointly across the entire fleet, with aircraft identity and age-related variables used as inputs. The goal is to evaluate how incorporating ageing information improves prediction accuracy and robustness.

\subsection{Baselines}

\textbf{Baseline 1 corresponds to a methodology based on the BADA framework \cite{nuic2010user} (hereafter referred to as BADA for simplicity)} — The EUROCONTROL BADA 4.2 model provides physics-based parametric fuel flow estimations widely used in trajectory simulation and emissions inventories. While highly interpretable and computationally efficient, BADA assumes nominal, time-invariant aircraft performance and does not model ageing or engine degradation effects. To compute the fuel flow we are using the BADA 4.2.1 model (via the \emph{pyBADA} library). Maximum thrust is assumed during the climb phase without any thrust derate and idle thrust is used during descent. True airspeed and vertical speed are smoothed using Savitzky--Golay filtering, and acceleration is then empirically derived from the smoothed true airspeed (TAS).

\textbf{Baseline 2: Acropole  \cite{jarrytowards}} — The model is a feed-forward neural network with multiple dense layers and optional dropout for regularization. All inputs are normalized and outputs are denormalized using fixed statistics to ensure consistency with physical units. L2 regularization is applied to all dense layers, and the final output is constrained to be non-negative via a ReLU activation.

\subsection{Ageing-Inclusive Modeling Approaches}

To address the limitations of the baselines, we propose and compare several methods to integrate aircraft age or engine deterioration effects into fuel flow prediction:

\textbf{BADA Seymour \cite{seymour2020fuel}} — A parametric correction based on a logarithmic deterioration coefficient proposed by Seymour et al.\ that scales baseline fuel consumption according to aircraft age, capturing the empirical fuel penalty due to ageing.

\textbf{BADA Calibrated} — An adaptation of the Seymour formula in which the deterioration coefficient is calibrated using observed QAR data and BADA prediction, allowing for improved fit and better representation of real-world ageing trends.

\textbf{Acropole Age} — A data-driven regression model built upon the Acropole baseline architecture, augmented by explicitly including aircraft age as an additional input feature to capture nonlinear interactions between ageing and other flight parameters.

\textbf{Acropole Inductive Bias} — Compared to the standard neural architecture that treats all input features identically, the proposed model introduces a structured parametrization to explicitly model the effect of aircraft age as an inductive bias. Both models employ input and output normalization, multiple dense layers with L2 regularization, and ReLU-constrained outputs. However, in this approach, the "age" feature is separated from the other normalized inputs before being fed into the neural network at the final layer. The output of the network, denoted as $\hat{y}_{\text{base}}$, is multiplied by an age-dependent coefficient 
$\text{Coeff}(\text{age}) = 1 + a \cdot \log(\text{age}+1)$, parametrized as a differentiable function of age. 
The final prediction is $\hat{y} = \max\!\left(0,\, \hat{y}_{\text{base}} \times \text{Coeff}(\text{age})\right)$.

This explicit multiplicative correction enforces the effect of ageing directly in the output, embedding domain knowledge about performance deterioration into the model structure, rather than relying solely on generic feature learning.

\subsection{Evaluation Metrics and Testing}

Models are trained on disjoint training, validation, and test sets, with final evaluation on the test set. Performance is measured with several metrics, where $\mathcal{D}$ is the dataset of pairs $(x,y)$ and $h$ the model:

\begin{itemize}[leftmargin=*]
    \item \textbf{MAE}: average absolute deviation,
    $\text{MAE}(h,\mathcal{D}) = \tfrac{1}{|\mathcal{D}|} \sum_{(x,y)} |h(x)-y|$.
    \item \textbf{MAPE}: relative error,
    $\text{MAPE}(h,\mathcal{D}) = \tfrac{1}{|\mathcal{D}|} \sum_{(x,y)} \tfrac{|h(x)-y|}{y}$.
    \item \textbf{ME}: signed bias,
    $\text{ME}(h,\mathcal{D}) = \tfrac{1}{|\mathcal{D}|} \sum_{(x,y)} (h(x)-y)$.
    \item \textbf{MSE}: squared deviation,
    $\text{MSE}(h,\mathcal{D}) = \tfrac{1}{|\mathcal{D}|} \sum_{(x,y)} (h(x)-y)^2$.
    \item \textbf{Bias ratio}: $\text{ME}^2/\text{MSE}$, fraction of error due to systematic bias.
\end{itemize}

\subsection{Training}

\noindent
The neural network models were implemented and trained using TensorFlow on a workstation equipped with two NVIDIA RTX 6000 GPUs. The model architecture comprised five hidden layers with 128 units per layer, ReLU activations, and L2 regularization ($\lambda = 1\times 10^{-5}$). Optimization was performed using AdamW with a warm-up cosine learning rate schedule, where the learning rate increased linearly during the first 10\% of the 200 training epochs, followed by a cosine decay towards zero. The batch size was selected to be 4096. Model performance was monitored on a held-out validation set using mean absolute error (MAE) and mean absolute percentage error (MAPE) metrics. Training checkpoints were saved automatically, with model parameters corresponding to the lowest validation loss retained for subsequent evaluation.

\section{Results}
\label{sec:result}

\begin{figure*}[h!]
    \centering

    \begin{subfigure}[b]{0.43\textwidth}
        \centering
        \includegraphics[width=\textwidth]{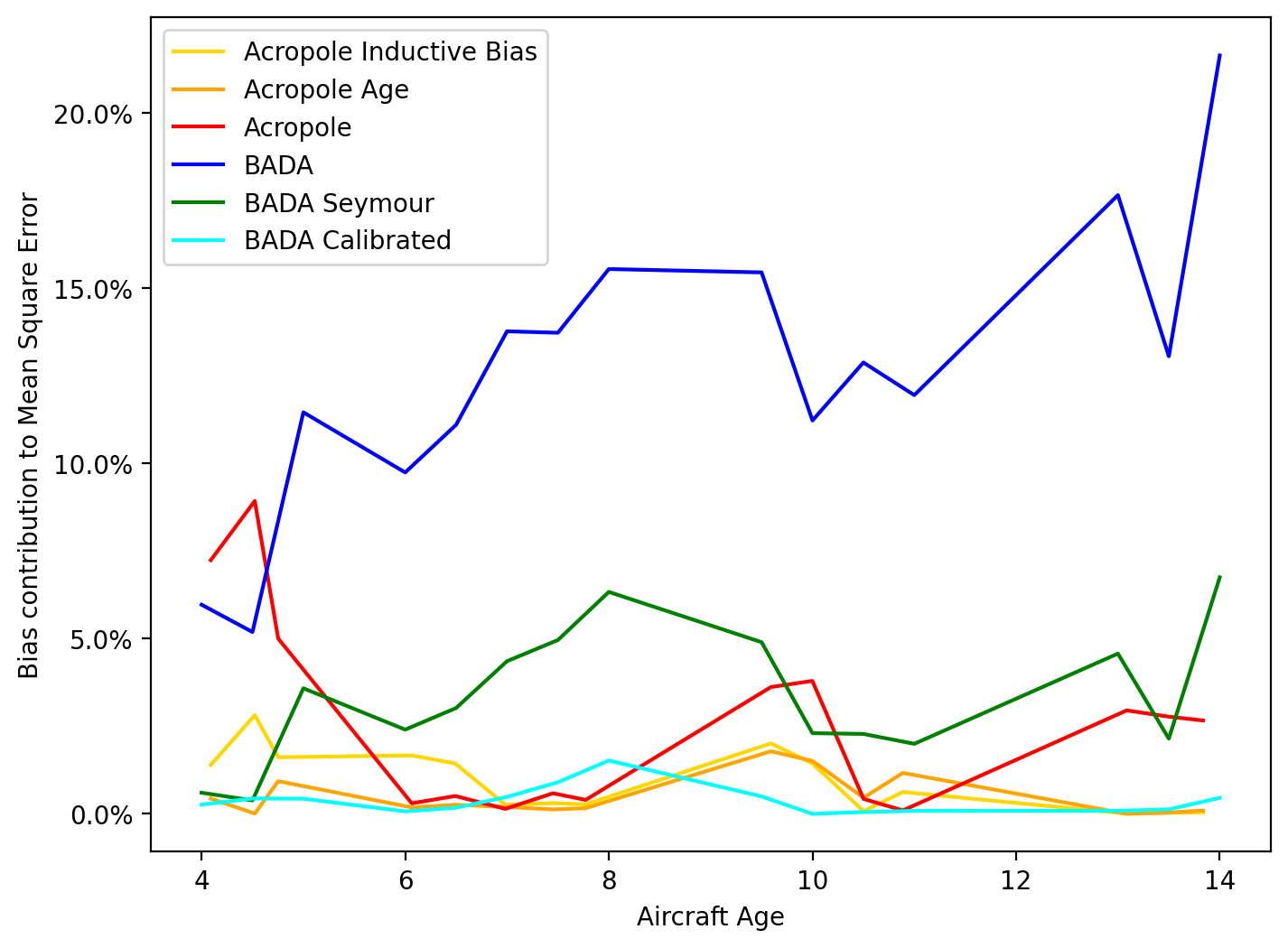}
        \caption{Bias contribution to Mean Squared Error (\%) vs. aircraft age}
        \label{fig:bias_to_mse}
    \end{subfigure}
    \hfill
    \begin{subfigure}[b]{0.43\textwidth}
        \centering
        \includegraphics[width=\textwidth]{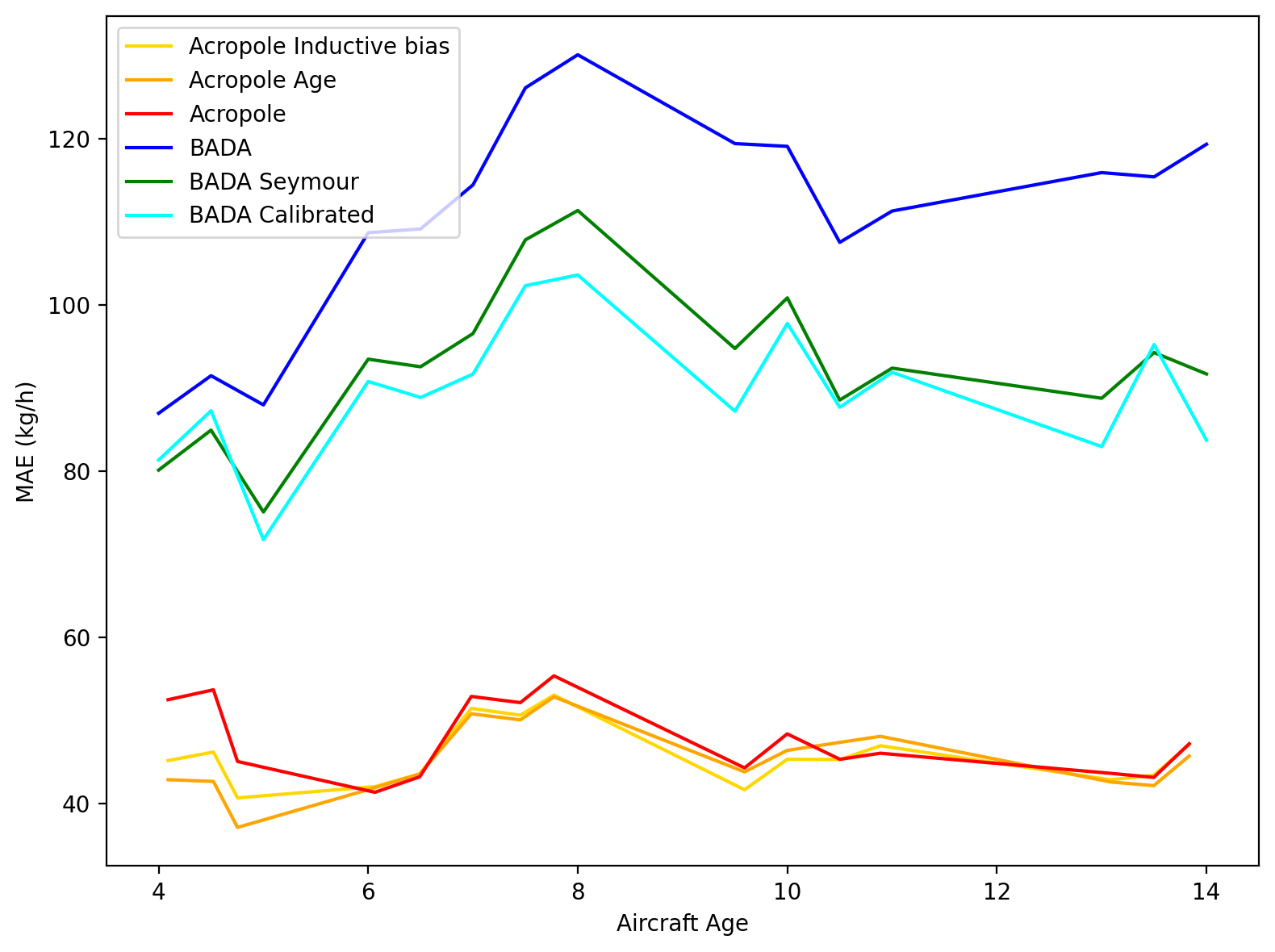}
        \caption{Mean Absolute Error (MAE) in kg/h vs. aircraft age}
        \label{fig:mae}
    \end{subfigure}

    \vspace{0.25cm}

    \begin{subfigure}[b]{0.43\textwidth}
        \centering
        \includegraphics[width=\textwidth]{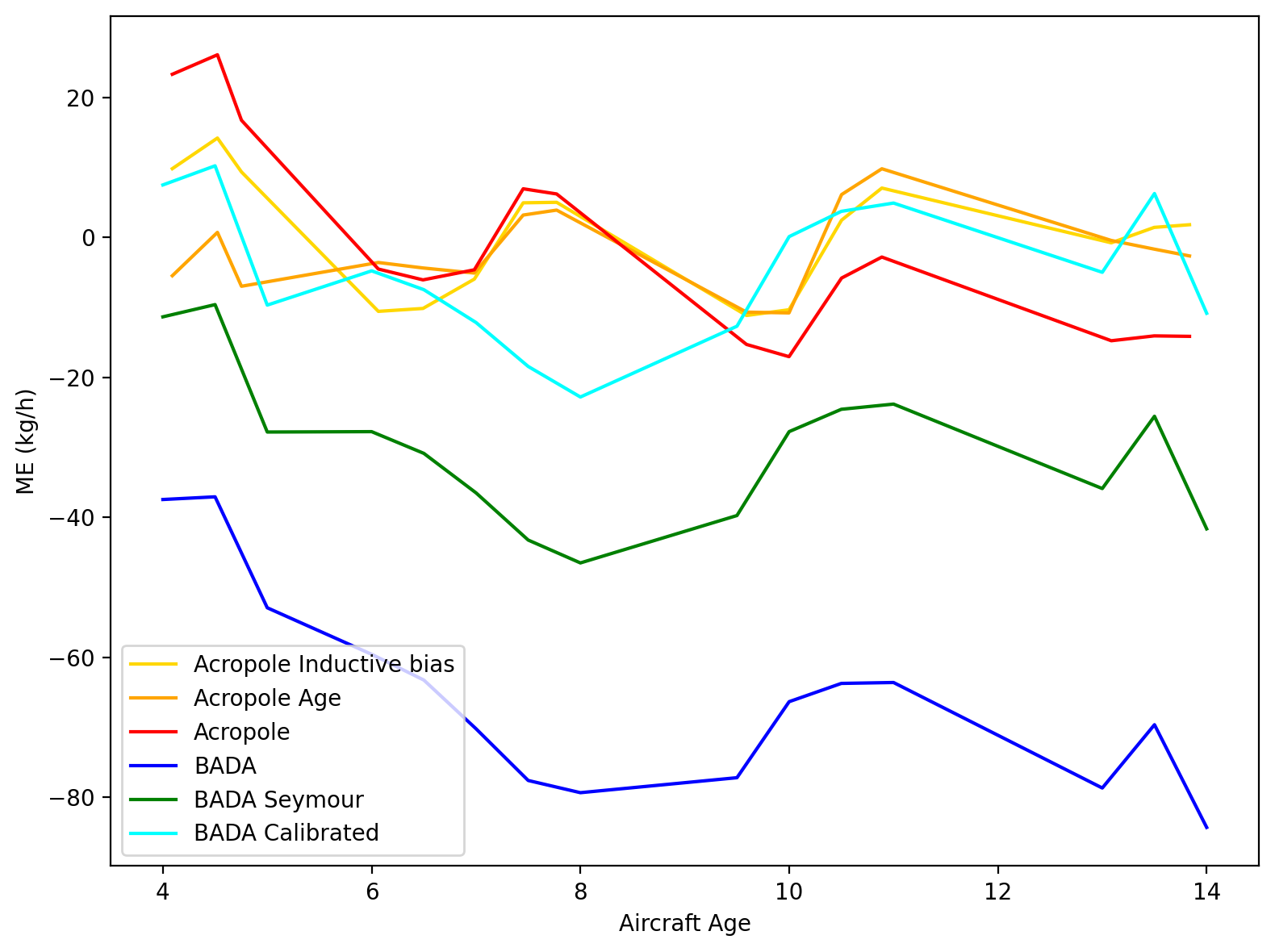}
        \caption{Mean Error (ME) in kg/h vs. aircraft age}
        \label{fig:me}
    \end{subfigure}
    \hfill
    \begin{subfigure}[b]{0.43\textwidth}
        \centering
        \includegraphics[width=\textwidth]{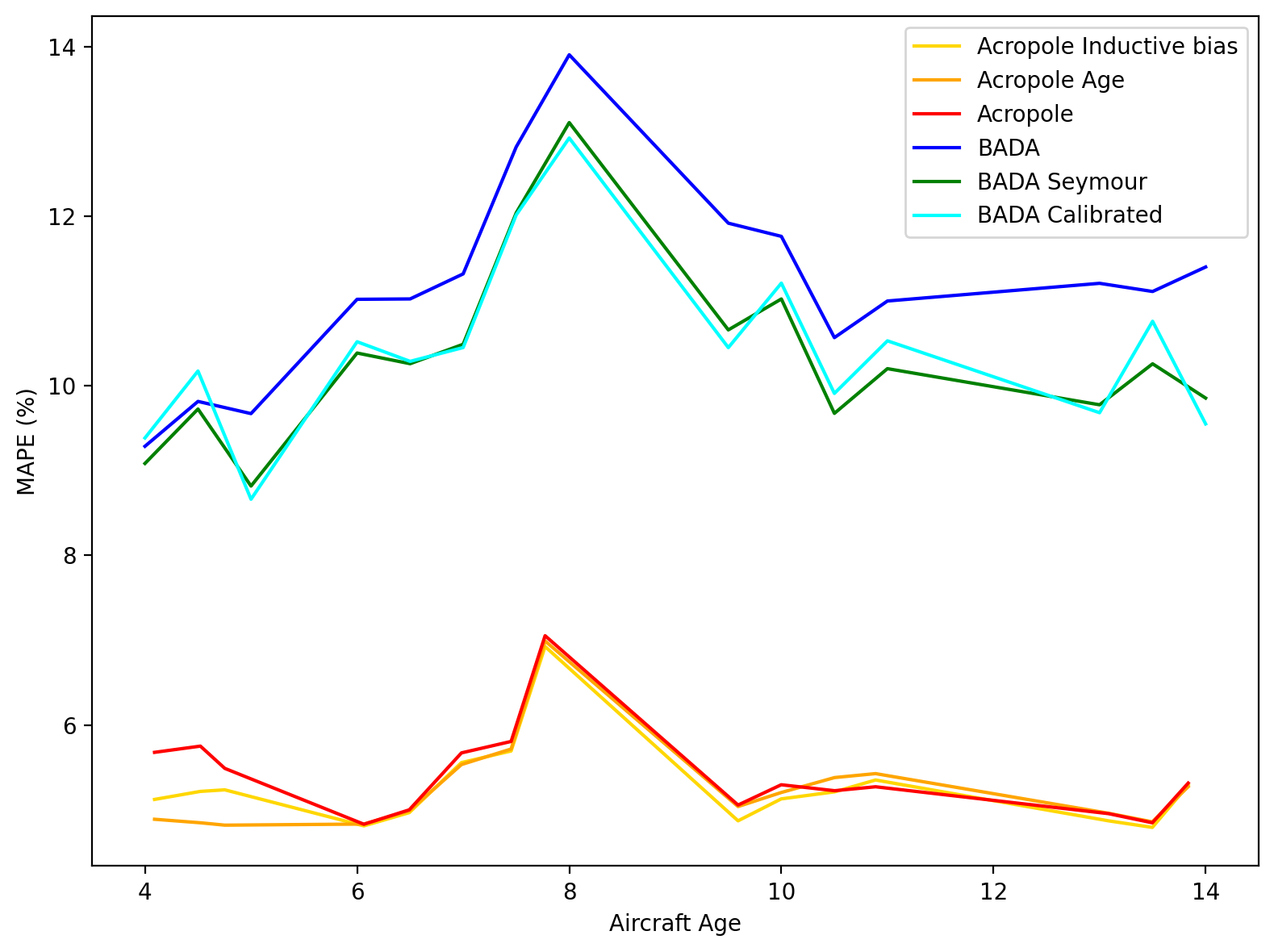}
        \caption{Mean Absolute Percentage Error (MAPE) (\%) vs. aircraft age}
        \label{fig:MAPE}
    \end{subfigure}

    \caption{Error metrics as a function of aircraft age for baselines and different modeling approaches}
    \label{fig:error_metrics_all}
\end{figure*}

\noindent
\subsection{Comparison of Baselines}

In this section we analyze the predictive tendencies of the baseline models and the impact of incorporating ageing corrections on fuel flow estimation accuracy.

The BADA model demonstrates a systematic bias that favors newer aircraft: specifically, it increasingly underestimates fuel consumption as aircraft age. This trend, illustrated by the blue line in Figures~\ref{fig:mae} and~\ref{fig:me}, is expected given that the model was originally trained on manufacturer data. As previously shown in~\cite{jarry2021toward}, the BADA model exhibits an average mean absolute error (MAE) of approximately $100\,\mathrm{kg/h}$ (Figure~\ref{fig:mae}) and an average underestimation bias of $50$--$60\,\mathrm{kg/h}$. Notably, the contribution of this bias to the mean squared error (MSE) ranges between $5\%$ and $20\%$ (Figure~\ref{fig:bias_to_mse}), indicating that the remaining error is primarily due to modeling limitations rather than age-related bias.

Incorporating the Seymour correction coefficient (green lines) reduces the bias contribution to the MSE to less than $5\%$, and to nearly $0\%$ when using the calibrated coefficient (cyan lines). Correspondingly, the direct bias decreases to $-30\,\mathrm{kg/h}$ with the Seymour correction and to approximately $-5\,\mathrm{kg/h}$ with calibration. It should be noted, however, that even with this age correction, the combined MAE remains around $85\,\mathrm{kg/h}$, which is still significantly higher than that observed with the Acropole model.

Analysis of the curves corresponding to the Acropole model (red lines) reveals that the model effectively learns an average fleet age. As a result, it tends to overestimate the fuel consumption of younger aircraft (by approximately $+20\,\mathrm{kg/h}$) and underestimate that of older aircraft (by approximately $-15\,\mathrm{kg/h}$), as shown in Figure~\ref{fig:mae}. Nevertheless, the bias contribution to the mean squared error (MSE) remains below $10\%$, with a mean absolute error (MAE) of about $45\,\mathrm{kg/h}$, indicating lower modeling error and enhanced robustness to age-related effects.

Incorporating age as an explicit input (orange lines) or as an inductive bias (gold line) further reduces the age-related bias to nearly zero. However, since the initial bias contribution was already below $5\%$, the resulting improvement in overall performance is moderate, with the average MAE decreasing only slightly to approximately $44\,\mathrm{kg/h}$. While the overall gain for the fleet is not substantial, it is noteworthy that the modeling of age is particularly beneficial for younger aircraft, for which the MAE decreases from $50\,\mathrm{kg/h}$ to $40\,\mathrm{kg/h}$.

\begin{figure}[ht]
    \centering
    \includegraphics[width=0.45\textwidth]{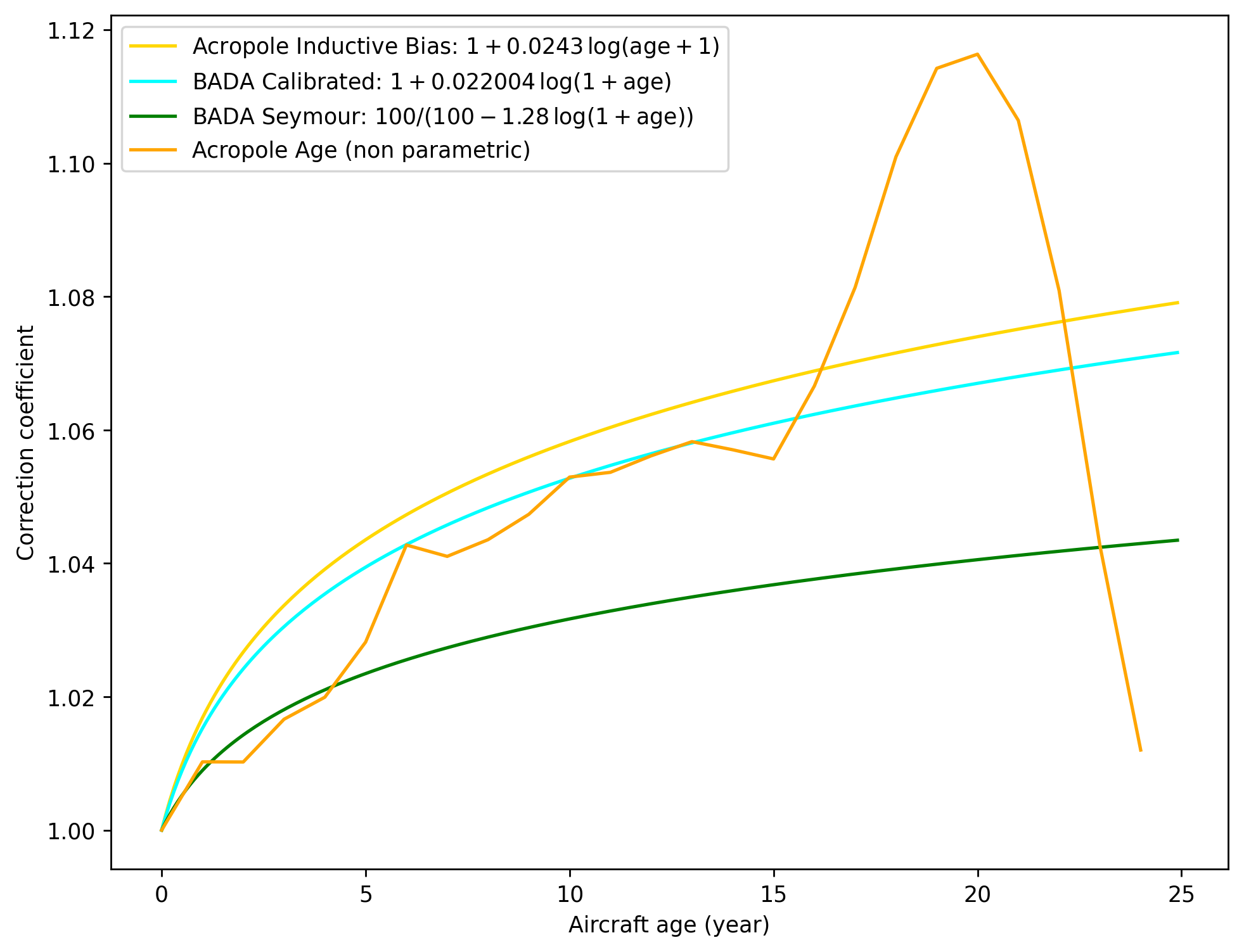}
    \caption{Correction coefficient function of aircraft age for the different modelling approaches}
    \label{fig:coeff}
\end{figure}

Additionally, Figure~\ref{fig:coeff} presents the evolution of the correction coefficients as a function of aircraft age. The coefficient for \textit{Acropole Age} (orange), which corresponds to a non-explicit internal value derived from the neural network, was estimated as follows: for each data point, the predicted fuel flow was first computed by setting the age to zero, and this baseline value was then used to normalize subsequent predictions for ages ranging from $0$ to $25$ years.

Several observations can be drawn from the correction coefficient curves. First, the Seymour coefficient appears to underestimate engine deterioration with increasing age. This may be attributed to its derivation from a dataset comprising different aircraft types and a limited number of trajectories. In contrast, the three other calibration methods exhibit consistent trends, showing an increase in fuel consumption of up to $6\%$ for aircraft at $20$ years of age. Notably, the \textit{Acropole Age} approach, where age is explicitly provided as an input, yields a less smooth correction curve that adapts closely to observed training values. However, since the dataset contains only nine distinct aircraft, this modelling approach is likely subject to overfitting due to limited representativeness—a limitation discussed further in Section~\ref{sec:discussion}. 

Moreover, for ages greater than $15$ years (for which no observed data are available), the model exhibits a non-monotonic increase in the correction coefficient then decrease, further highlighting that, in situations with limited data representativeness, the introduction of inductive bias can provide more robust model characteristics.

\subsection{Impact on Fuel Consumption Assessment}

\subsubsection{Current fleet impact}
\begin{table*}[h!]
\centering
\caption{Computed consumption over test set in Tonnes of fuel}
\label{tab:data}
\begin{tabular}{ccccccccc}
\toprule
& Ground Thruth & BADA & BADA Seymour & BADA Calibrated & Acropole & Acropole Inductive Bias & Acropole Age \\
Consumption (tonnes) & 23176.4 & 22061.9 & 22673.5 & 23115.2 & 23124.8 & 23142.2 & 23141.2 \\
Difference (tonnes) &    & -1114.5 & -502.9   & -61.2   & -51.6   & -34.3   & -35.2   \\
Difference Ratio &    & -4.81\%  & -2.17\%  & -0.26\%  & -0.22\%  & -0.15\%  & -0.15\%  \\
\bottomrule
\end{tabular}
\end{table*}

To evaluate the practical implications of the different modelling approaches on environmental impact assessments, we computed the total fuel consumption over the entire test dataset for each model. This cumulative consumption provides insight into how each model's prediction differences translate into estimated fuel use at fleet scale, a key factor in operational planning and emissions accounting.

Table~\ref{tab:data} summarizes the total fuel consumption (in tonnes) predicted by each model alongside the ground truth measured consumption. The raw BADA model underestimates total fuel burn by approximately 4.81\% (1,115 tonnes) compared to observed values, reflecting its nominal performance assumption without ageing effects. Incorporating the Seymour parametric correction reduces this underestimation by more than half, to around 2.17\% (503 tonnes), while the calibrated BADA variant further narrows the gap to a marginal 0.26\% (61 tonnes).

The data-driven Acropole models provide even closer estimates, with differences below 0.22\% (52 tonnes) across all three age-aware variants. Among these, the Acropole Inductive Bias approach achieves the smallest discrepancy of 0.15\% (34 tonnes), indicating the benefit of explicitly embedding ageing effects into the model structure.

These results highlight that failing to account for engine deterioration and ageing in fuel flow models can lead to significant underestimation of fleet fuel consumption and associated emissions. Conversely, integrating age-dependent corrections or data-driven ageing representations materially improves fuel burn estimates, enhancing the reliability of environmental impact assessments. This has important ramifications for regulatory compliance, airline operational planning, and long-term sustainability strategies.

While the overall absolute differences may appear modest relative to total fuel use, even small percentage errors translate to substantial quantities of fuel and CO$_2$ emissions when aggregated over the scale of global commercial aviation. Therefore, precise modelling of ageing effects is critical to underpin robust policy and operational decisions aimed at mitigating aviation’s environmental footprint.

\subsubsection{Current fleet in the future}

To evaluate the long-term impact of engine ageing on fleet fuel consumption, we simulate cumulative fuel use over a horizon of 1 to 15 years, assuming a constant fleet size and no new aircraft produced. The simulated fleet corresponds to the nine A320-214 airframes in our dataset, whose ages at the start of the simulation range from 3 to 14 years (average age 8 years). 

The baseline reference consumption curve was defined as the average of the correction coefficients obtained from the two best-performing models in our study, namely the \textit{BADA Calibrated} and the \textit{Acropole Inductive Bias} approaches. Since the true functional form of engine degradation with age is not directly observable, we cannot rely on a single ``ground truth'' curve. Instead, we approximate the underlying trend by combining the results of these two complementary models. This averaging ensures a more robust baseline, as it reduces the influence of model-specific biases and reflects the central tendency of the most accurate methods available. The resulting function, $1 + 0.0231 \log(1 + \text{age})$,
therefore represents our best estimate of the expected long-term increase in fuel consumption due to ageing effects at the fleet level.

\begin{figure}[ht]
    \centering
    \includegraphics[width=0.45\textwidth]{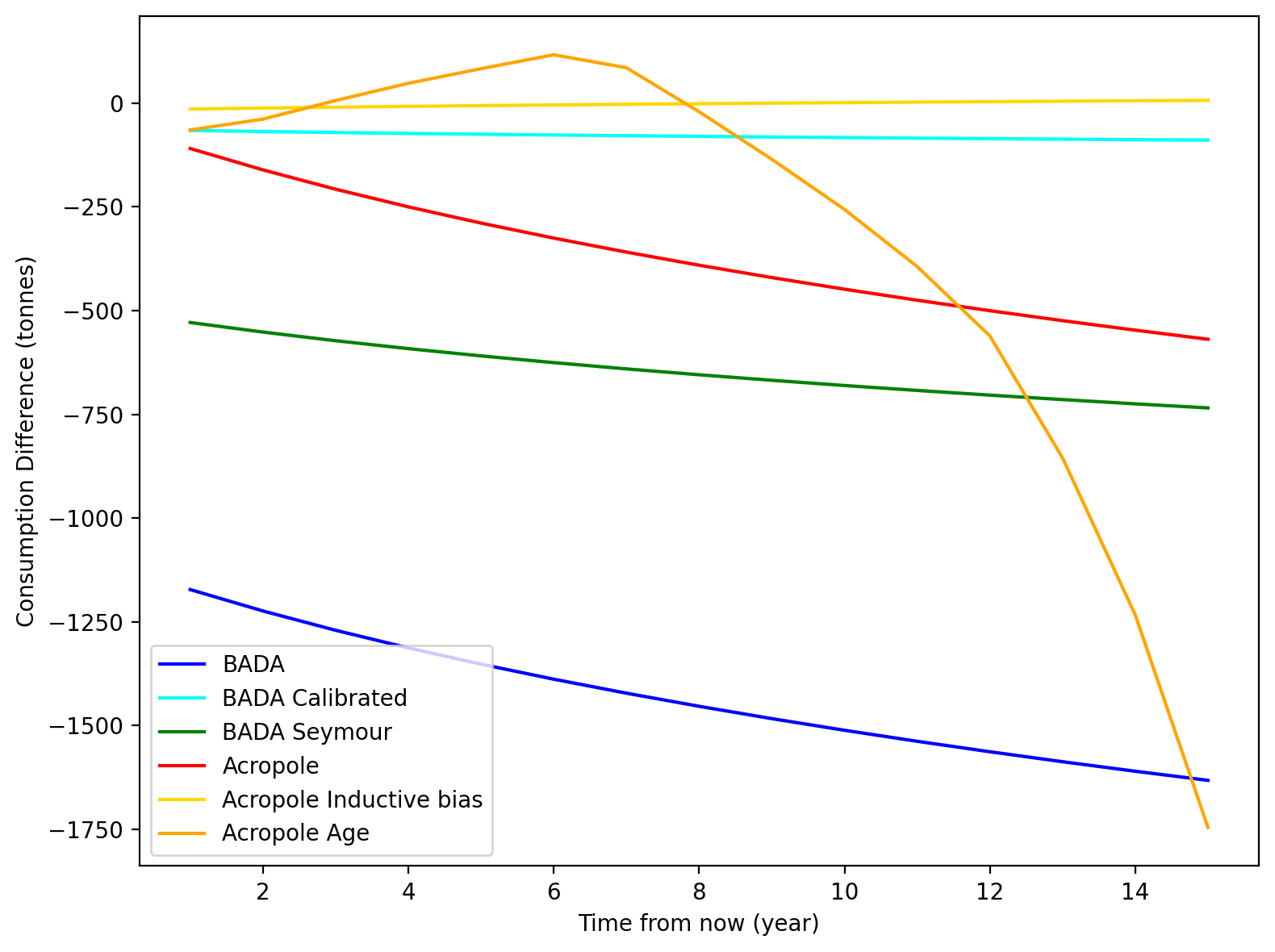}
    \caption{Projected fuel consumption differences (tonnes) over a 1–15 year horizon under the assumption of a constant fleet.}
    \label{fig:simu}
\end{figure}

Figure~\ref{fig:simu} presents the projected differences in fuel consumption (in tonnes) for each modelling approach relative to this baseline scenario.

The simulation results show that the raw BADA model substantially underestimates the additional fuel burn compared to the reference deterioration baseline, with the discrepancy increasing from approximately 1,172 tonnes at 1 year to over 1,600 tonnes at 15 years. These figures correspond to the nine-aircraft fleet considered in this study, i.e. an average deviation of about 130–180 tonnes per aircraft over the 15-year horizon. A similar trend is observed for the BADA Seymour correction, albeit with smaller deviations.

Data-driven models exhibit more varied behaviour: the Acropole model slightly overestimates consumption increases during the initial years, gradually converging towards the BADA Seymour predictions. The Inductive Bias model and BADA Calibrated model demonstrate smaller errors and minor divergence, consistent with their averaging in the chosen baseline correction coefficient.

Finally, the Acropole Age model displays more oscillatory behaviour and, as previously discussed, tends to substantially under-predict fuel consumption once the average fleet age exceeds the maximum age observed during training.

These projections highlight the sensitivity of long-term fuel consumption forecasts to the selection of ageing models and calibration methodologies. Despite the limited representativeness of the training dataset, the introduction of inductive bias appears to enhance model robustness. Overall, this analysis confirms that incorporating age-related corrections is essential and is best achieved using calibrated coefficients.

\section{Discussion}
\label{sec:discussion}

A key limitation of this study is the restricted number of individual airframes available, with only nine aircraft of the same type. This narrow sampling constrains the representativeness of the age parameter, particularly when attempting to model the complex relationship between fuel consumption and aircraft ageing using fully data-driven neural network approaches. To partially mitigate this, the dataset was split for each aircraft into two temporal subsets based on calendar dates, effectively doubling the temporal coverage and improving the model’s ability to capture ageing trends within the limited fleet. Nevertheless, this approach cannot fully compensate for the limited number of distinct airframes.

In real operational fleets, maintenance events such as engine washes, overhauls, or component replacements periodically reset the effective engine degradation, producing non-monotonic patterns of performance loss that are not captured when age is measured solely as time since entry into service. Without access to detailed maintenance or cleaning cycle histories, our models may conflate true age-related deterioration with aircraft-specific idiosyncrasies, leading to overfitting and poor generalization outside the observed age range.

As demonstrated in the fleet projection analysis (Section~\ref{sec:result}), the choice of ageing model significantly affects long-term fuel consumption forecasts. Under the assumption of a constant fleet, the raw BADA based methodology tends to substantially underestimate additional fuel burn, while calibrated and inductive bias approaches yield more consistent and realistic predictions. The observed oscillations in the Acropole Age model further underscore the risks of extrapolating beyond the training data without robust inductive constraints.

Consequently, while our age-parameterized neural network models can capture ageing trends present in the available data, their ability to generalize and robustly represent ageing effects across a more diverse fleet remains limited if no inductive bias is introduced. Expanding the dataset to include a greater number of aircraft, ideally with detailed records of engine cleaning and maintenance events, would be essential to improve both the accuracy and generalizability of age-based corrections in data-driven fuel flow models. Introducing inductive bias in modelling ageing effects appears to be a promising direction to mitigate data scarcity and improve model robustness.

\section{Conclusions}
\label{sec:conclusions}

This study has provided a systematic evaluation of methods for incorporating ageing and engine deterioration effects into aircraft fuel-flow modelling. Using a large QAR dataset covering nine A320-214 airframes with a wide range of service ages, we have shown that standard parametric models, such as BADA based methodology, systematically underestimate fuel consumption for older aircraft. The application of empirical correction coefficients from the literature, as well as coefficients calibrated on observed data, substantially reduces bias.

Neural network approaches that explicitly model aircraft age, either as an additional input feature or through an inductive bias parameterization, offer improved performance, especially in reducing mean error and enhancing robustness to within-fleet variability. However, the limited number of distinct airframes constrains the representativeness of age and can result in overfitting or limited generalization. The absence of maintenance history data, such as cleaning or engine overhaul events, further limits the ability to distinguish true age-related effects from aircraft-specific variations.

The simulation of future fleet fuel consumption under a constant fleet size assumption reveals that models without age correction may substantially misestimate long-term fuel burn, potentially biasing operational and environmental analyses. Models incorporating calibrated correction coefficients and inductive biases provide more realistic projections, emphasizing the importance of robust ageing representations.

Future work should focus on expanding datasets to include broader and more diverse fleets, ideally with detailed maintenance records, to strengthen the reliability and applicability of age-based corrections in data-driven fuel-flow models. 

\balance 
\bibliographystyle{IEEEtran}
\bibliography{references}

\end{document}